\documentclass[journal]{IEEEtran}

\usepackage{lineno,hyperref}
\usepackage{amsthm}
\usepackage{amsfonts}
\modulolinenumbers[5]
\usepackage{amsmath,multirow,url,fancyhdr}
\usepackage{amsmath}
\usepackage{algorithm, algorithmic}

\def\1{\mathbf{1}}

\def\N{\mathcal{N}}
\def\I{\mathcal{I}}

\def\S{{\mathcal S}}

\def\k{{\mathbf k}}

\def\X{{\bf X}}
\def\x{{\bf x}}

\def\y{{\bf y}}
\def\D{{\mathbf D}}

\def\real{{\mathbb R}}
\newcommand{\norm}[1]{\|{#1}\|}

\newtheorem{theorem}{Theorem}[section]
\newtheorem{remark}{Remark}[section]

\ifCLASSINFOpdf
 \usepackage[pdftex]{graphicx}
\else
\fi
\usepackage{algorithmic}

\begin{document}
%
\title{Sequential Randomized Matrix Factorization for Gaussian Processes: Efficient Predictions and Hyper-parameter Optimization}
%
%
%

\author{Shaunak D. Bopardikar,~\IEEEmembership{Member,~IEEE}
     and  George S. Eskander Ekladious
\thanks{Shaunak D. Bopardikar and George S. Eskander Ekladious are with the United Technologies Research Center, East Hartford, Connecticut,
 06108 USA e-mail: bopardsd@utrc.utc.com, ekladigs@utrc.utc.com.}
}

\maketitle

\begin{abstract}
This paper presents a sequential randomized low-rank matrix factorization approach for incrementally predicting values of an unknown function at test points using the Gaussian Processes framework. It is well-known that in the Gaussian processes framework, the computational bottlenecks are the inversion of the (regularized) kernel matrix and the computation of the hyper-parameters defining the kernel. The main contributions of this paper are two-fold. First, we formalize an approach to compute the inverse of the kernel matrix using randomized matrix factorization algorithms in a streaming scenario, i.e., data is generated incrementally over time. The metrics of accuracy and computational efficiency of the proposed method are compared against  a batch approach based on use of randomized matrix factorization and an existing streaming approach based on approximating the Gaussian process by a finite set of basis vectors. Second, we extend the sequential factorization approach to a class of kernel functions for which the hyperparameters can be efficiently optimized. All results are demonstrated on two publicly available datasets.
\end{abstract}


%
\IEEEpeerreviewmaketitle

\section{Introduction}

Gaussian processes is a powerful framework for several machine learning tasks such as regression, classification and inference. Given a finite set of input output training data that is generated out of a fixed  (but possibly unknown) function, the framework models the unknown function as a stochastic process such that the training outputs are a finite number of jointly Gaussian random variables, whose properties can then be used to infer the statistics (the mean and variance) of the function at test values of input. The computation can be implemented in a \emph{batch} setting, i.e., one-shot over the entire training data, or in a \emph{sequential} setting where the data is processed incrementally. In either setting, the scalability of the computation grows with the number of points in the training data. The accuracy of the predictions depends upon the values of a set of parameters that define the Gaussian process, termed as the \emph{hyperparameters}. The hyperparameters are selected by optimizing an appropriate cost function that captures the prediction error over the dataset, and is typically a complex problem from a computational perspective. This paper addresses both of these scalability aspects of Gaussian processes -- a randomized matrix factorization approach to efficiently compute the predictions and an efficient approach for the hyperparameter selection problem presently applicable only to a class of kernel functions. 

\subsection{Related Literature}

The framework of Gaussian processes and their application to regression, classification, etc. is well established. The  reader is referred to \cite{rasmussen2006} and \cite{alvarez2012} for a detailed review of the methodology. The computational complexity of Gaussian processes is centered around the need to invert the \emph{kernel matrix} whose size is governed by the number of training data points. Therefore, the worst-case computational complexity of the method is $O(n^3)$, where $n$ is the number of training points. Over the past decade, several approaches have been proposed to improve the scalability using sparse approximations (\cite{candela2005}, \cite{hensman2013}). Representing the Gaussian processes using a set of $k \ll n$ basis vectors, the computational complexity can be lowered down to $O(k^2n)$, but in a batch setting \cite{smola2001, williams2000, snelson2005}. \cite{banerjee2013} applied randomized matrix factorization techniques to compute low rank approximations of the kernel matrix. Recently, \cite{oneill2016} showed that for some classes of kernel functions, the covariance matrix can be hierarchically factored into a product of block low-rank updates of the identity matrix, thereby leading to an $O(n\log^2(n))$ algorithm for inversion. Although the method applies to several commonly used kernels, to the best of our knowledge, the accuracy of the method has not been quantified to kernels outside of that class.

In the context of hyperparameter optimization, several techniques based on sparse representations of Gaussian Processes have been addressed in literature \cite{smola2001, williams2000, snelson2005}. These technique choose a subset of $k$ vectors by optimizing some information criterion. Specifically, \cite{snelson2005} finds the basis vector locations and the hyper-parameters in a single smooth joint optimization formulation. More recently, \cite{cao2013} presented an efficient discrete optimization algorithm to jointly estimate a sparsity inducing subset of training data and to choose the hyperparameters. \cite{huber2014} developed recursive approaches to Gaussian processes based on updating the basis vectors set efficiently at every time step followed by a step based on learning/updating the hyperparameters. The complexity of their basic approach scales as $O(bk^2)$, where $b$ is the size of the new samples at every time step, and thus does not increase over time/iterations.

Computations for streaming applications have received a lot of recent attention. \cite{liberty2013} developed a simple deterministic approach to derive a \emph{sketch} (a much smaller representative approximation) of any matrix. \cite{ghashami2015} and \cite{huang2015} apply concepts from randomized matrix sketching to address streaming versions of Principal Components Analysis. These works bear some similarity with the proposed framework with respect to the eigenvalue problems that are required to be solved, but the techniques differ in the iterative usage of the projection based approach that we leverage. This paper is an expanded version of the preliminary study on some of the ideas that are proposed in \cite{Bopardikar2016}, where the Gaussian process is updated sequentially, but \emph{without optimizing the hyperparameters}. 

\subsection{Contributions}

We present a sequential randomized low-rank matrix factorization approach for incrementally predicting values of an unknown function at test points using the Gaussian Processes framework. We propose an approach to compute the inverse using randomized matrix factorization algorithms in a streaming scenario, i.e., data is generated incrementally over time. The use of randomized matrix factorization allows us to decide the computational complexity by selecting a fixed \emph{target rank} for the factorization. The metrics of accuracy (factorization error) and computational efficiency of the proposed methods are compared against a batch approach based on use of randomized matrix factorization on publicly available real datasets. We also provide a comparison of the accuracy of the proposed method in the mean squared sense with the method from \cite{huber2014} on publicly available real datasets.

We then introduce a new technique for optimizing the hyperparameters. While most techniques rely on sparsification of the Gaussian Process and then jointly optimize the hyperparameters, our technique is based on a sequential update of a low-rank factorization of the kernel matrix in a way that the kernel matrix and its inverse can be quickly re-computed for hyperparameter optimization. This method currently works only for a selected class of kernel functions which are affine in the hyperparameters and the pairwise Euclidean distance matrix of the training dataset. The results are reported on publicly available datasets.

The distinguishing contributions of this paper as compared to the early work \cite{Bopardikar2016} are: 1) the hyperparameter optimization approach for a class of kernel functions; 2) theoretical bound on the approximation accuracy and the computational complexity of the proposed algorithms, and 3) a more detailed comparative study by testing the proposed algorithms against a state-of-the-art batch random matrix factorization algorithm.
 
\subsection{Organization of the paper}
Section~\ref{sec:problem} provides a brief description of Gaussian processes, Randomized Matrix Factorization techniques and the problem being addressed. Section~\ref{sec:sequential} summarizes the proposed Sequential Random Matrix Factorization (SMRF) approach.  Section~\ref{sec:Incremental Gaussian Processes} describes how the proposed approach is applied to the incremental Gaussian process regression problem and presents the hyperparameter optimization approach.  Section~\ref{sec:Simulation Results} presents numerical results on benchmark data for comparison. Finally, Section~\ref{sec:conclusion} summarizes the work done with some directions identified for future research.

\section{Problem Formulation and Background} \label{sec:problem}
In this section, we review some basic concepts related to Gaussian processes from the review \cite{alvarez2012} as well as randomized algorithms for matrix factorization from \cite{NH-PGM-JAT:11}.
\subsection{Gaussian Processes (GP): Prediction}
In the classical setting of supervised learning, given a training set $\S := (\X,\y) = (\x_1, y_1), \dots , (\x_n, y_n)$, one of the most popular estimators of the output at a new point $\x_{n+1}$ can be defined through a Bayesian perspective. In this context, a Gaussian process (GP) is a stochastic process with the property that any finite number of random variables taken from a realization of the GP, follows a joint Gaussian distribution. If the underlying function $f$ follows a Gaussian process, we write
\[
f \sim GP(\mu, K),
\]
where $\mu$ is the mean function and $K$ the covariance or kernel function. The mean function and the covariance
function completely specify the Gaussian process. In other words, for any finite
set $\X = \{\x_1,\dots, \x_n\}$, if we let $f(\X) = [f(\x_1), \dots , f(\x_n)]^T$, then
\[
f(\X) \sim \N(\mu(\X), K(\X,\X)),
\]

where $\mu(\X) = [\mu(\x_1), \dots ,\mu(\x_n)]^T$ and $K(\X,\X)$ is the kernel matrix which is symmetric positive semidefinite with $K(\x_i, \x_j) = h(\x_i, \x_j, \phi)$ where $\phi$ is a set of parameters defining the kernel. Without any loss of generality, we assume that the mean vector is zero.
In a regression context, the likelihood function is usually Gaussian and expresses a linear relation between the
observations and a given model for the data that is corrupted with a zero mean Gaussian noise,
\[
p(y|f, x, \sigma^2) = \N(f(x), \sigma^2),
\]
where $\sigma^2$ corresponds to the variance of the noise. Noise is assumed to be independent and identically distributed. For a test input vector $\x_{n+1}$, the posterior distribution is given by
\[
p(f(\x_{n+1})|\S, \x_{n+1},\phi) = \N(f_*(\x_{n+1}), K_*(\x_{n+1}, \x_{n+1})),
\]
where
\begin{align}\label{eq:gp}
f_*(\x_{n+1}) &= \k_{\x_{n+1}}^T(K(\X,\X) + \sigma^2I)^{-1}\y, \nonumber \\
K_*(\x_{n+1},\x_{n+1}) &= K(\x_{n+1},\x_{n+1}) \nonumber \\
&- \k_{\x_{n+1}}^T(\underbrace{K(\X,\X) + \sigma^2I}_{\I})^{-1} \k_{\x_{n+1}}.
\end{align}
and $\k_{\x_{n+1}}^T:=\begin{bmatrix}K(\x_1,\x_{n+1}) & \dots & K(\x_n,\x_{n+1})\end{bmatrix}^T$, is the vector of covariances of the test input vector $\x_{n+1}$ with the training set.

\subsection{Inversion of positive semi-definite kernel matrices}\label{sec:inversion}
Recall that any positive semi-definite covariance matrix $K$ can be spectrally decomposed as $K = USU^T$, where $U$ is the matrix of eigenvectors such that $U^TU = I$ and $S$ is a diagonal matrix with the entries equal to the eigenvalues. Then, the Woodbury identity is usually used to compute the inverse of a positive definite matrix of the type
\begin{align*}
X^{-1} &= (A + BCD)^{-1} \\ &= A^{-1} - A^{-1}B(C^{-1} + DA^{-1}B)^{-1}DA^{-1}.
\end{align*} 
Applying this identity to the matrix $\I$, we obtain
\begin{align}\label{eq:Iinv}
\I^{-1} &= \sigma^{-2}I - \sigma^{-4}U(S^{-1} + \sigma^{-2}I)^{-1}U^T.
\end{align}

\subsection{Randomized Spectral Decomposition for Positive Definite Matrices}

Given any matrix $A \in \real^{m\times n}$, a target rank $k$ and an oversampling parameter $p$, Algorithm~\ref{algo:fixedrank}  returns a matrix $Q \in \real^{m\times (k+p)}$ with orthonormal columns such that
\begin{equation}\label{eq:mf}
\norm{A - QQ^TA} \leq \Big( 1 +  9\sqrt{k+p}\sqrt{\min\{m,n\}} \Big) \sigma_{k+1}(A),
\end{equation}
with probability at least $1 - 3p^{-p}$ and where the norm $\norm{\cdot}$ denotes the spectral two norm.




\begin{algorithm}[t]
\begin{algorithmic}[1]
\STATE \textbf{Input:}  $A \in \real^{m\times n}$, $0 < k < \min\{m,n\}$, 
and a positive integer $p$.
    \STATE {\bfseries Output:} Matrix $Q$ such that $\lVert{A-QQ^TA}\rVert\in O(\sigma_{k+1}(A))$
with probability at least $1 - 3p^{-p}$.
\STATE Draw a random $n \times (k+p)$ test matrix $\Omega$.
\STATE Form the matrix product $Y = A\Omega$.
\STATE Orthonormalize columns of $Y$ and set them equal to $Q$.
\end{algorithmic}
\caption{Fixed rank Factorization (Algo. 4.1 from \cite{NH-PGM-JAT:11}) } \label{algo:fixedrank}
\end{algorithm} 

Further, given any positive definite matrix $A \in \real^{n\times n}$ and a matrix $Q$ such that \eqref{eq:mf} holds, Algorithm~\ref{algo:nystrom} yields an approximate eigenvalue decomposition of $A$.

\begin{algorithm}[t]
\begin{algorithmic}[1]
\STATE \textbf{Input:} Positive semi-definite $A$.
\STATE \textbf{Output:} $U, S$ such that $A \approx U S U^T$.
\STATE Compute $Q$ using Algorithm~\ref{algo:fixedrank}. 
\STATE Form the small matrix $B = Q^TAQ$.
\STATE Compute the eigenvalue decomposition $B = V S V^T$. 
\STATE Set $U = QV$.
\end{algorithmic}
\caption{Approx. Eigen-decomposition (Algo. 5.1 \cite{NH-PGM-JAT:11}). } \label{algo:nystrom}
\end{algorithm} 

The computational complexity of Algorithm~\ref{algo:nystrom} is $O(kn^2 + k^2n)$, where the dominant cost is that of computation of the matrix $Q$. The goal of this paper is to develop a streaming implementation for Gaussian process prediction, with
complexity $O(n)$ per iteration.

\section{Sequential Randomized Kernel Matrix Factorization}\label{sec:sequential}

The central idea behind improving the computational complexity for the sequential/streaming application is to use the decomposition $A \approx U S U^T$ obtained at iteration $t$ to compute the corresponding decomposition at iteration $t+1$. Let the kernel matrix at iteration $t+1$ be given by
\[
 K_{t+1} := \begin{bmatrix} A & B \\ B^T & C \end{bmatrix},
\]
where $B, C$ can be matrices of appropriate dimensions. Now, to compute the factorization $K_{t+1} \approx \bar{U} \bar{S} \bar{U}^T$, we will use the matrix
\[
\bar{K}_{t+1} := \begin{bmatrix} USU^T & B \\ B^T & C \end{bmatrix},
\]
in place of $K_{t+1}$. The action of matrix $K_{t+1}$ on a random test vector $\omega^{(i)}$, which is computationally the most expensive step (step 4 in Algorithm~\ref{algo:fixedrank}), is now substituted by 

\begin{align}\label{eq:gp2}
y^{(i)} = \begin{bmatrix} USU^T & B \\ B^T & C \end{bmatrix}\begin{bmatrix} \omega_1 \\ \omega_2 \end{bmatrix}
\end{align}
where the random vector $\omega^{(i)} = [\omega_1; \omega_2]$, has been appropriately partitioned. Algorithm~\ref{algo:sequential} summarizes the sequential approach.

\begin{algorithm}[t]
\begin{algorithmic}[1]
\STATE \textbf{Initialization:} Compute $U_1, S_1$ such that $K_1 = U_1S_1U_1^T$, using either exact or approximate method.
\FOR{$t \in \{2, 3, \dots\}$}{
\STATE Compute $Q_{t}$, using Algorithm~\ref{algo:fixedrank} with positive parameters $p$ and $k$, where $k<n$, for the matrix
\[
\bar{K}_{t} := \begin{bmatrix} U_{t-1}S_{t-1}U_{t-1}^T & B \\ B^T & C \end{bmatrix},
\]
where matrices $B, C$ are the new data at iteration $t$.
\STATE Form the small matrix $Z = Q_{t}^T\bar{K}_{t}Q_{t}$.
\STATE Compute the eigenvalue decomposition $Z = VS_t V^T$. 
\STATE Set $U_{t} = Q_tV$.}
\ENDFOR
\end{algorithmic}
\caption{Sequential Approximate Eigen-decomposition. } \label{algo:sequential}
\end{algorithm} 

The following Theorem summarizes the accuracy of Algorithm~\ref{algo:sequential}.

\bigskip
\begin{theorem}\label{thm:sequential}
For every $T$, after the $T$-th iteration of Algorithm~\ref{algo:sequential}, the approximation error 
\[
\norm{K_T - U_TS_TU_T^T} \leq 2\sum_{t=1}^T \Big( 1 +  9\sqrt{(k+p)n_t} \Big) \sigma_{k+1}(K_t),
\]
with probability at least $1 - 3Tp^{-p}$, where $p$ is the oversampling parameter from Algorithm~\ref{algo:fixedrank} and $n_t$ is the dimension of the kernel matrix $K_t$ at the $t$-th iteration. 
\end{theorem}

\bigskip

Although the right hand side is a sum of $T$ terms, in the event that $k$ is chosen such that each of the $\sigma_{k+1}(K_t)$ is small, the error remains small. Further, the error also holds with a high probability since even for a small choice of $p$, the number of iterations $T$ until which the error bound holds can be sufficiently high $O(p^p)$.

The proof is based on repeated application of the accuracy guarantee on Algorithm~\ref{algo:fixedrank} from~\cite{NH-PGM-JAT:11}, and is similar in spirit to the bound established in \cite{bopardikar2013} for updating the factorization for generic matrices. The main difference is in the specific structure in positive definite kernel matrices. 

\medskip

\begin{proof}
 For brevity, let $\epsilon_k := \Big( 1 +  9\sqrt{(k+p)n_k} \Big) \sigma_{k+1}(K_k)$. We begin with
\begin{align*}
&\norm{K_T - U_TS_TU_T^T} \leq \norm{K_T - \bar{K}_T} + \norm{\bar{K}_T - U_TS_TU_T^T} \\
&= \norm{\begin{bmatrix} K_{T-1} & B_T \\ B_T^T & C_T \end{bmatrix} - \begin{bmatrix} U_{T-1}S_{T-1}U_{T-1}^T & B_T \\ B_T^T & C_T \end{bmatrix}} \\&+  \norm{\bar{K}_T - U_TS_TU_T^T} \\
&= \norm{K_{T-1} - U_{T-1}S_{T-1}U_{T-1}^T} + \norm{\bar{K}_T - U_TS_TU_T^T} \\
& \vdots \\
&\leq \sum_{t = 1}^T \norm{\bar{K}_t - U_tS_tU_t^T}.
\end{align*}
Now, $\bar{K}_t$ is symmetric, and if $Q_t$ is a matrix with orthonormal columns such that $ \norm{\bar{K}_t- Q_tQ_t^T\bar{K}_t} \leq \epsilon_t$, then we obtain that
\begin{align*}
&\norm{\bar{K}_t - Q_tQ_t^T\bar{K}_tQ_tQ_t^T} \\
&= \norm{\bar{K}_t- Q_tQ_t^T\bar{K}_t + Q_tQ_t^T\bar{K}_t  - Q_tQ_t^T\bar{K}_tQ_tQ_t^T} \\
&\leq \norm{\bar{K}_t- Q_tQ_t^T\bar{K}_t} + \norm{Q_tQ_t^T(\bar{K}_t  - \bar{K}_tQ_tQ_t^T)} \\
&\leq 2\epsilon_t.
\end{align*}
where the final step follows from the fact that $\norm{Q_tQ_t^T} \le 1$.

Now, let $\mathcal{E}_t$ denote the event that $\norm{\bar{K}_t- Q_tQ_t^T\bar{K}_t} \leq 2\epsilon_t$. Then,
\begin{align*}
\mathbb{P}(\bigcap_{t=1}^T \mathcal{E}_t) = 1 - \mathbb{P}((\bigcap_{t=1}^T \mathcal{E}_t)^c) &= 1 - \mathbb{P}(\bigcup_{t=1}^T \mathcal{E}_t^c) \\
&\geq 1 - \sum_{t=1}^T\mathbb{P}(\mathcal{E}_t^c). 
\end{align*}
Now, from \eqref{eq:mf}, we have that $\mathbb{P}(\mathcal{E}_t^c) \leq 3p^{-p}n_t$. Since the event $\mathcal{E}_t$ implies that $\norm{\bar{K}_t - U_tS_tU_t^T} \leq 2\epsilon_t$, the proof is complete.
\end{proof}

\bigskip

\begin{remark}[Computational complexity]
The main advantage of using the proposed approach over batch one is that the computational bottleneck of matrix-matrix multiplication in  step 4 of Algorithm~\ref{algo:fixedrank} is reduced from $O(n^2k)$ down to $O(nk^2)$ per iteration.

\end{remark}

\section{Incremental Gaussian Processes}\label{sec:Incremental Gaussian Processes}
In this section, we present the application of the sequential randomized matrix factorization approach (Sequential Approximate Eigen-decomposition Algorithm~\ref{algo:sequential}) to Gaussian processes in the streaming scenario.
We first introduce a generic algorithm that applies to any kernel and that does not involve hyperparameter optimization (Algorithm~\ref{algo:seqGP}). Then, we introduce an algorithm for incremental GP with hyperparameter optimization that applies to a specific class of kernels with the property of having the hyperparameters easily decoupled from the pairwise distance matrix. This property enables optimizing the hyperparameters in a sequential way. More specifically, we derive the method with simple quadratic kernels (Algorithm~\ref{algo:seqGPhyper}), but the method can be generalized to any polynomial kernel.

\subsection{Without Hyperparameter Optimization}\label{sec:without-optimization}

The technique is summarized in Algorithm~\ref{algo:seqGP}. Any valid kernel function can be selected for this technique. The hyperparameters defining the kernel function are kept constant at every iteration. The initial step (\# 2) can be computed using either exact or approximate method. At every subsequent iteration, the low-rank factorization of the kernel matrix is updated (\# 4-9), which is then used to compute the inverse (\# 10). The outputs are then predicted using the closed form expressions for Gaussian processes (\#11).



\begin{algorithm}[t]
\begin{algorithmic}[1]
\STATE   \textbf{Initialization:} At $t=1$, compute $K_1$ using a Kernel function and the initial  input $x_1$.
\STATE  {Compute $U_1, S_1$ such that $K_1 = U_1S_1U_1^T$,  using either exact or approximate method.}

\FOR{$t \in \{2, 3, \dots\}$}{
\STATE Compute cross covariances $\k_{\x_{t}}^T:=\begin{bmatrix}K(\x_1,\x_{t}) & \dots & K(\x_{t-1},\x_{t})\end{bmatrix}^T)$, and $K(\x_{t},\x_{t}) $ using $\x_t$.
\STATE Form
\[
\bar{K}_{t} := \begin{bmatrix} U_{t-1}S_{t-1}U_{t-1}^T & \k_{\x_{t}}  \\ \k_{\x_{t}} ^T & K(\x_{t},\x_{t}) \end{bmatrix},
\]

\STATE Compute $Q_{t}$ for the matrix $\bar{K}_{t}$ using Algorithm~\ref{algo:fixedrank} where  \eqref{eq:gp2}  is applied in step 4,  with positive parameters $p$ and $k<b$, where $b$ is  the size of the new arrived data $x_t$.

\STATE Form the small matrix $Z = Q_{t}^T\bar{K}_{t}Q_{t}$.
\STATE Compute the eigenvalue decomposition $Z = VS_t V^T$.
\STATE Set $U_{t} = Q_tV$.
\STATE Compute $\I^{-1}= \sigma^{-2}I - \sigma^{-4}U_t(S_t^{-1} + \sigma^{-2}I)^{-1}U_t^T$}
\STATE Predict outputs for the new input $\x_t$ using \eqref{eq:gp}.
\ENDFOR
\end{algorithmic}
\caption{Sequential Randomized GP Regression} \label{algo:seqGP}
\end{algorithm} 

This technique relies on a good choice of the hyperparameters, which could often be very adhoc.  On the other hand, it is not possible to optimize the hyperparameters  by employing the  sequential matrix factorization process  listed in Algorithm \ref{algo:seqGP}.
The reason  is  that the kernel factorization is being computed recursively using the stored factored form of the previous kernel instance. Accordingly, it is not possible to measure the impact of a new candidate set of hyperparameters, on the GP prediction accuracy, since the change in  the  hyperparameters will not  affect the stored part of the kernel, which has been computed using the old hyperparameters.
   
 The next sub-section addresses this problem and provides an efficient way to learn and update the hyperparameters in a sequential mode for a specific class of kernel functions.

\subsection{With Hyperparameter Optimization}\label{sec:with-optimization}

We introduce the following kernel function. Given a set of input vectors $\X := \{\x_1, \dots, \x_n\}$ and a vector of  hyperparameters $\mathbf{a} := [a_0 \dots a_{m-1}]$ (where $a$ must have non-negative enteries), for a positive integer $m$, define
\begin{equation}\label{eq:K}
K_\mathbf{a}(\X, \X) := a_0I + \sum_{i=1}^{m-1}a_i {\D^i(\X,\X)};
\end{equation}
where $I$ is the $n\times n$ identity matrix, $\D(\X, \X)$ is the pair-wise Euclidean distance matrix between the input vectors and the operator ${\D^i(\X,\X)}$ will yield an $n \times n$ matrix in which every element is the corresponding element in $\D$ raised to $i$. In other words, for $i=2$, ${\D^2(\X,\X)} = \D \circ \D$ is the Hadamard product of $\D$ with itself, and in general, ${\D^i(\X,\X)} = \underbrace{\D \circ \ldots \circ \D}_{i \text{ times}}$. From Schur's theorem \cite{schur1911}[Theorem VII], we have that the Hadamard product of positive semi-definite matrices is positive semi-definite and therefore, ${\D^i(\X,\X)}$ is positive semi-definite for every $i \in \{2,\dots, m-1\}$. Finally, notice that $K_\mathbf{a}$ is a sum of positive semi-definite matrices and therefore, is positive semi-definite and a valid kernel. 

We will now describe our approach for the hyperparameter optimization. The key idea here is to maintain and update a low rank factorization of ${\D^i(\X,\X)}$, for every $i$. i.e., ${\D^i(\X,\X)} \approx U_iS_iU_i^T$, which in turn will lead to an efficient computation as well as the update of $K$ in the following manner. Suppose that $m=2$ for illustrative purposes. Then,
\begin{align*}
K  &= a_0 I + a_1 \D + a_2 {\D^2(\X,\X)}\\
&\approx a_0 I + a_1 U_1S_1U_1^T + a_2 U_2S_2U_2^T. \\ 
\end{align*}
We can write $\underbrace{K + \sigma^2I}_{\I}=  P + a_2 U_2S_2U_2^T$, where \\ $P := (a_0+\sigma^2) I + a_1 U_1S_1U_1^T$.  Applying the Woodbury identity (Section~\ref{sec:inversion}) to $P$, we have
\begin{equation}\label{eq:Pinv}
P^{-1} = \frac{1}{a_0+\sigma^2} I - \frac{1}{ { (a_0+ \sigma^2)}^2}U_1\big ( \frac{1}{a_1}S_1^{-1} + \frac{1}{a_0+\sigma^2} I \big )^{-1} U_1^T.
\end{equation}
Once we have $P^{-1}$, we can apply the Woodbury identity to the above  expression of $K$ to obtain
\begin{align}\label{eq:Kinv}
K^{-1} &\approx P^{-1} - P^{-1}U_2\Big ( \frac{1}{a_2}S_2^{-1} + U_2^TP^{-1}U_2 \Big)^{-1}U_2^T P^{-1}.
\end{align}

\begin{remark}[Computational complexity]
The computational complexity of this approach when we use Algorithm~\ref{algo:sequential} to update the factors of $\D$ and $\D^2$ at every iteration is $O(pnk^2)$, assuming that the matrices $U_1, U_2$ have at most $k$ columns each. This follows from the facts that
\begin{enumerate}
\item Updating the factorization of $\D$ and in general of $\D^i, \forall i \in \{2,\dots, m\}$, is $O(nk^2)$.  
\item Computation of $P^{-1}$ is $O(nk^2)$ since it is dominated by computing the product in the second term of (6).
\item The computation of the term $U_2^TP^{-1}U_2$ is $O(k^3)$ since one can use the expression for $P^{-1}$ to first compute $U_2^TU_1$, resulting into the product of three $k\times k$ matrices. 
\item Finally the second term in the expression for $K^{-1}$ can be computed in $O(nk^2)$ steps as it is a product of one $n\times k$ matrix $P^{-1}U_2$, a $k\times k$ matrix (central term) and a $k\times n$ matrix.
\end{enumerate}
\end{remark}

\bigskip
The key insight behind this approach is that one does not need to re-compute the kernel matrix from scratch every time the hyperparameters are updated during the optimization step. Due to the \emph{separable} structure of the terms involving the distance matrix and the hyperparameters, one can quickly recompute the kernel matrix (and its inverse) at every iteration.  The complete approach is summarized in Algorithm~\ref{algo:seqGPhyper}.

\begin{algorithm}[t]
\begin{algorithmic}[1]
\STATE   \textbf{Initialization:} At $t=1$, compute $\D^i, \forall i \in \{1,\dots, m-1\}$.
\STATE  {Compute factors $U_i, S_i$ such that $\D^i = U_iS_iU_i^T$, using either exact or approximate method.}

\FOR{$t \in \{2, 3, \dots\}$}{

\STATE Compute  pairwise distances  ${\D_{\x_{t}}}^T:=\begin{bmatrix}D(\x_1,\x_{t}) & \dots & D(\x_{t-1},\x_{t})\end{bmatrix}^T)$, and $D(\x_{t},\x_{t}) $ using $\x_t$.

\STATE Form
\[
\bar{D}_{t} := \begin{bmatrix} U_{t-1}S_{t-1}U_{t-1}^T & \D_{\x_{t}}  \\ \D_{\x_{t}} ^T & D(\x_{t},\x_{t}) \end{bmatrix}.
\]

\STATE Compute $Q_{t}$ for the matrix $\bar{D}_{t}$ using Algorithm~\ref{algo:fixedrank} where  \eqref{eq:gp2}  is applied in step 4,  with positive parameters $p$ and $k<b$, where $b$ is  the size of the new arrived data $x_t$.
\STATE Form the small matrix $Z = Q_{t}^T\bar{D}_{t}Q_{t}$.
\STATE Compute the eigenvalue decomposition $Z = VS_t V^T$.
\STATE Set $U_{t} = Q_tV$.
\STATE Repeat step 5-9 $\forall i \in \{1,\dots, m-1\}$.

\STATE Compute $K^{-1}_t$ using \eqref{eq:Pinv} and \eqref{eq:Kinv}.
\STATE Predict outputs for the training data $[x_1, \dots x_{t-1}]$ using \eqref{eq:gp}.

\STATE Update the hyperparameters to minimize prediction error for the training data (under the constraint that the hyperparameters vector [vector $a$ in \ref{eq:K} ]  has non-negative entries).
\STATE Repeat step 11-13 until convergence .
\STATE Compute $K^{-1}_t$ using \eqref{eq:Pinv} and \eqref{eq:Kinv} and the optimized hyperparameters.

\STATE Predict outputs for the new input $\x_t$ using \eqref{eq:gp}.}
\ENDFOR
\end{algorithmic}
\caption{Sequential GP with hyperparameter optimization} \label{algo:seqGPhyper}
\end{algorithm}




It now remains to be seen how this choice of kernel functions performs on real datasets. This will be addressed in the next section, specifically in \ref{subsec:With_Hyperparameter_Optimization}.

\section{Simulation Results}\label{sec:Simulation Results}
 In this section, we present numerical results of applying our techniques from Section~\ref{sec:Incremental Gaussian Processes} on synthetic as well as real datasets. We begin with the SRMF algorithm \ref{algo:seqGP} on two real world datasets. Accuracy and efficiency of the incremental Gaussian Process are compared against a baseline where No Random Matrix Factorization (NRMF) is applied, as well as the batch approach (BRMF). Also, the proposed approach is compared to a state-of-the-art Recursive Gaussian Process (RGP) algorithm where computational complexity is reduced by selecting a set of basis vectors \cite{huber2014}. Finally, applicability and limitations of the hyperparameter optimization approach, listed in Algorithm \ref{algo:seqGPhyper}, are discussed.

\subsection{Test Datasets}\label{sec:Results on Real World Datasets}

Two datasets are used for the comparative study.  The first  is the \emph{Abalone} dataset, from the UCI machine learning database, where the  age of abalone  is  predicted  given other eight attributes \cite{FRANK2010}.  For implementing the sequential learning scenario, only the first 4000 samples of the training set  are used. 

The second dataset is the \emph{Sarcos Robot arm} dataset which is publicly available at \cite{sarcos}. The data relates to an inverse dynamics problem for a seven degrees-of-freedom SARCOS anthropomorphic robot arm. The task is to map from a 21-dimensional input space (7 joint positions, 7 joint velocities, 7 joint accelerations) to the corresponding 7 joint torques. In our simulations, we predict the torque on the first joint (the $22^{nd}$ column) using the full input space (remaining 21 columns). The first 10,000 samples of the training dataset are used for simulations.

For both datasets, the batch size is set to 100 samples. We  simulate the case where  only a single batch of labeled data is available for training at the initial time, while all subsequent batches are unlabeled data that are sequentially sent to the GP for prediction. For this scenario, the GP is updated at each iteration using the predicted outputs since no actual outputs are available. This is a challenging problem since the prediction error might be accumulated  overtime and introduced as  noise to the GP.

\subsection{Without Hyperparameter Optimization}\label{sec:Without Hyperparameter Optimization}
We first evaluate Algorithm \ref{algo:seqGP} where a squared exponential kernel is employed for the covariance computations. The hyperparameters were empirically estimated and the same values are used for all the algorithms included in this comparative study. We compare the efficiency and regression accuracy when the incremental GP is driven by either of the random matrix factorization approaches (BRMF and SRMF) against a baseline GP, where no random matrix factorization is applied (NRMF), and the inverse of the full matrix is computed and used for regression. Also the random matrix factorization approaches are compared to the RGP method \cite{huber2014}.

\begin{figure}[!h]
\centering
\includegraphics[width=2.5in]{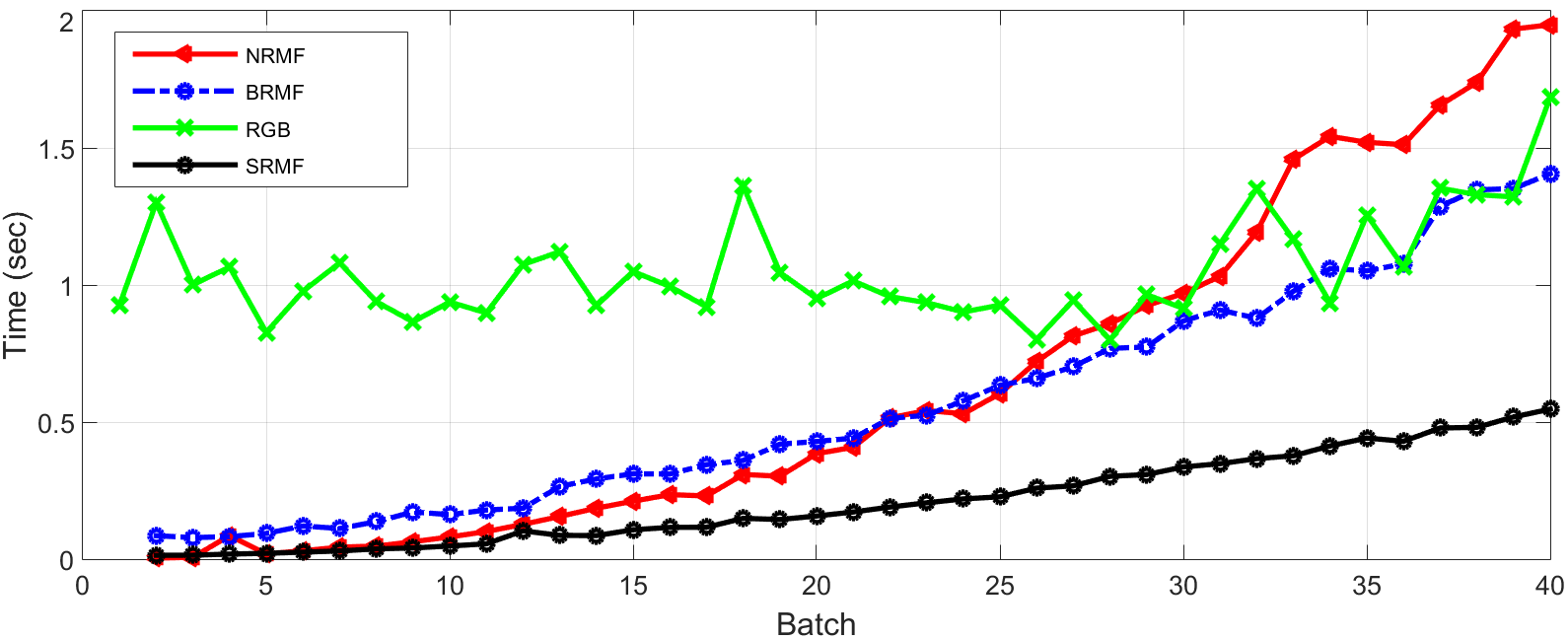}
\caption{Computational time for the Abalone dataset using the exponential kernel without hyperparameter optimization}
\label{fig:time_Abalone_expk}
\end{figure}

\begin{figure}[!h]
\centering
\includegraphics[width=\columnwidth]{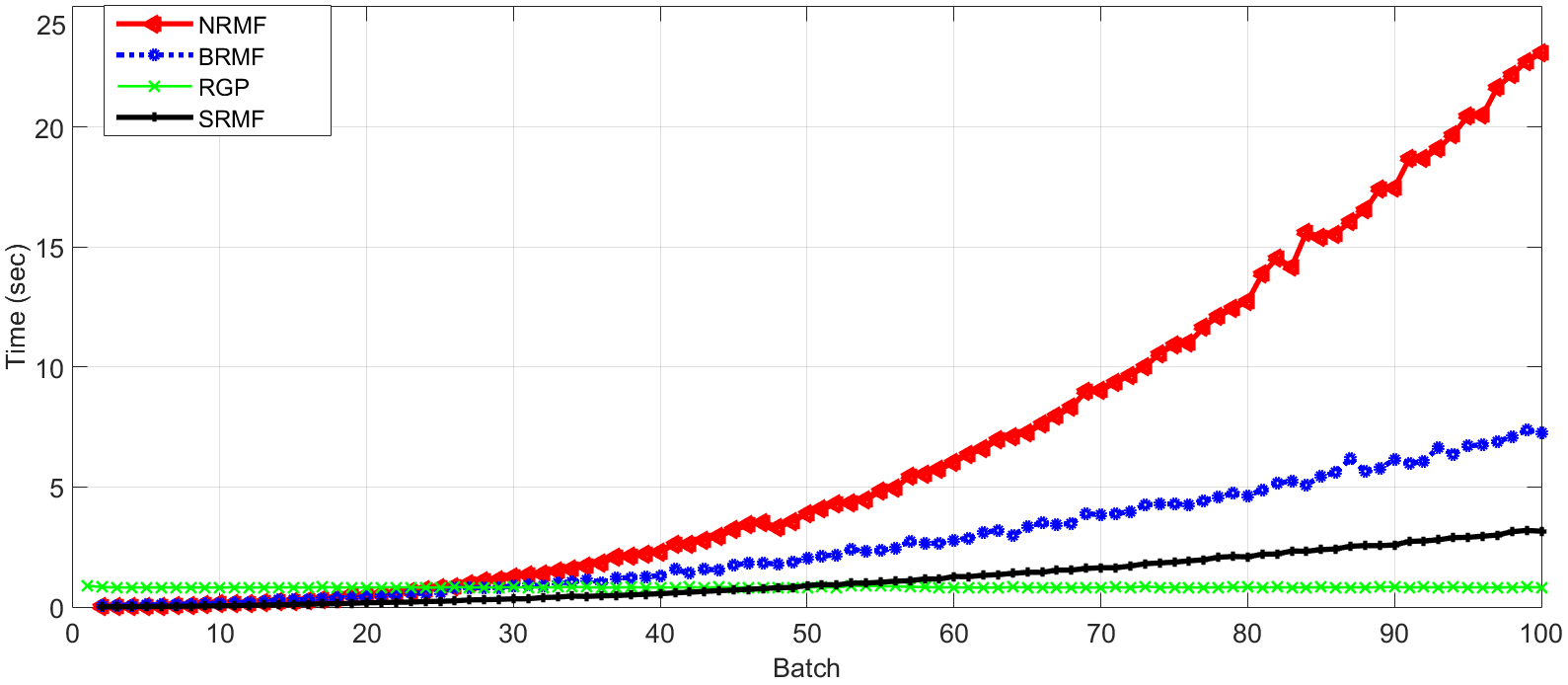}
\caption{Computational time for the Sarcos dataset using the exponential kernel without hyperparameter optimization}
\label{fig:time_Sarcos_expk}
\end{figure}

Figures~\ref{fig:time_Abalone_expk} and~\ref{fig:time_Sarcos_expk} show the computational time taken  for the  Abalone and Sarcos datasets, respectively. It is clear that the incremental GP scenario is inefficient when no factorization step is applied (NRMF). Both random factorization algorithms increase the efficiency, wherein the proposed sequential algorithm (SRMF) outperforms the batch algorithm (BRMF). 

While the recursive GP algorithm (RGP) has no significant computational variation, the proposed SRMF algorithm shows increasing computational time when a new batch updates the GP. During initial iterations, the proposed method is faster since the data size is small. The matrix factorization approach results in faster computations than the deterministic approach employed in the RGP algorithm. Although the deterministic approach keeps small fixed data size, it does not employ any step to factor the matrix for reduced sizes. Over increasing number of iterations, the proposed approach becomes slower since even the factored matrices become larger when data is accumulated. At some point, our algorithm will get slower than the RGP algorithm. For the Abalone simulations, our algorithm was faster for all iterations (Figure~\ref{fig:time_Abalone_expk}).  On the other hand, for the Sarcos simulations, the RGP shows a subtle better speed by the final iterations (Figure~\ref{fig:time_Sarcos_expk}).

\begin{table}[!t]

\caption{Comparison of Algorithms for Incremental GP with Exponential Kernel AND  Without Hyperparameter Optimization}
\centering
\begin{tabular}{|c|c|c|c|c|}
\hline
\multirow{2}{*}{Algorithm}& \multicolumn{2}{|c|}{Abalone}  & \multicolumn{2}{|c|}{Sarcos}\\\cline{2-5}
 {}& RMSE & Time & RMSE& TIME\\
\hline
NRMF& 2.73 & 0.63 &14.40& 6.68\\
\hline
RGP&  2.96 & 1.05 &14.80& 0.82\\
\hline
BRMF& 3.22 & 0.55 &8.88& 2.65\\
\hline
SRMF&  3.22 &  0.21 &8.88& 1.15\\
\hline
\end{tabular}
\label{table:expKNRMF}
\end{table}

Table \ref{table:expKNRMF} summarizes the comparative study where the exponential kernel is employed and the hyperparameters are manually set. Generally, the proposed algorithm significantly decreases the GP computational complexity and makes it feasible for sequential or streaming applications, while keeping the prediction accuracy at a similar/comparable level. Here we report the root mean square error (RMSE) and time averages over all batches (40 and 100 batches for the Abalone and Sarcos datasets. respectively). Generally, the  matrix factorization algorithms (BRMF and SRMF) provide a good compromise between accuracy and efficiency. It is clear that efficiency of the proposed sequential method (SRMF) significantly and consistently outperforms the batch algorithm for both of the two dataset simulations, without loss in the accuracy. 

For the Abalone dataset, employing the matrix factorization approach significantly increased the efficiency, when compared with the baseline with no matrix factorization applied (NRMF), with a subtle decrease in accuracy. However, for the Sarcos dataset, surprisingly, the accuracy is enhanced by employing the matrix factorization approaches. The factorization method introduces some noise to the GP, and this might be beneficial to model the uncertainties in data and distance metrics.  

Compared the RGP algorithm, the proposed SRMF seems to be more efficient for smaller datasets (e.g., Abalone), while it is less efficient for bigger datasets. However, the proposed SRMF algorithm shows enhanced accuracy for the Sarcos dataset and comparable accuracy for the Abalone dataset. 



\subsection{With Hyperparameter Optimization} \label{subsec:With_Hyperparameter_Optimization}

\begin{table}[!t]
\caption{Comparison of Optimization Modes for Incremental GP with Polynomial Kernel AND  Without Random Matrix Factorization}
\centering
\begin{tabular}{|c|c|c|c|c|}
\hline
\multirow{2}{*}{Optimization mode}& \multicolumn{2}{|c|}{Abalone}  & \multicolumn{2}{|c|}{Sarcos}\\\cline{2-5}
 {}& RMSE & Time & RMSE& TIME\\
\hline
No optimization& 55.60 & 0.60  &386.68& 6.62\\
\hline
Continous optimization& 3.76 & 125.77 &3.52&  595.72\\
\hline
Initial optimization& 5.06 & 1.72 &5.66& 6.86\\
\hline
\end{tabular}
\label{table:polyKNRMF}
\end{table}

\begin{figure}[!t]
\centering
\includegraphics[width=\columnwidth]{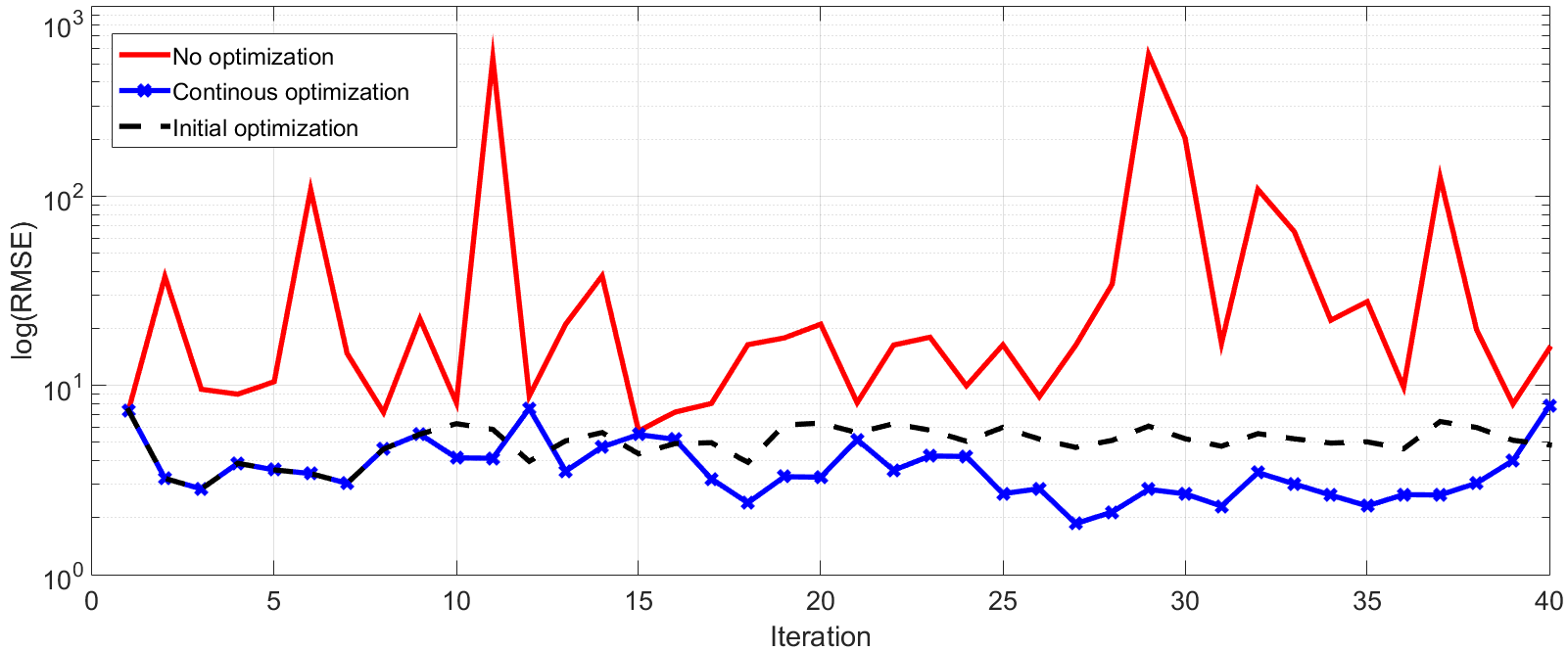}
\caption{Accuracy for the Abalone dataset using the polynomial kernel with hyperparameter optimization and without Random Matrix Factorization}
\label{fig:REMSE_Abalone_polyk}
\end{figure}

\begin{figure}[!t]
\centering
\includegraphics[width=\columnwidth]{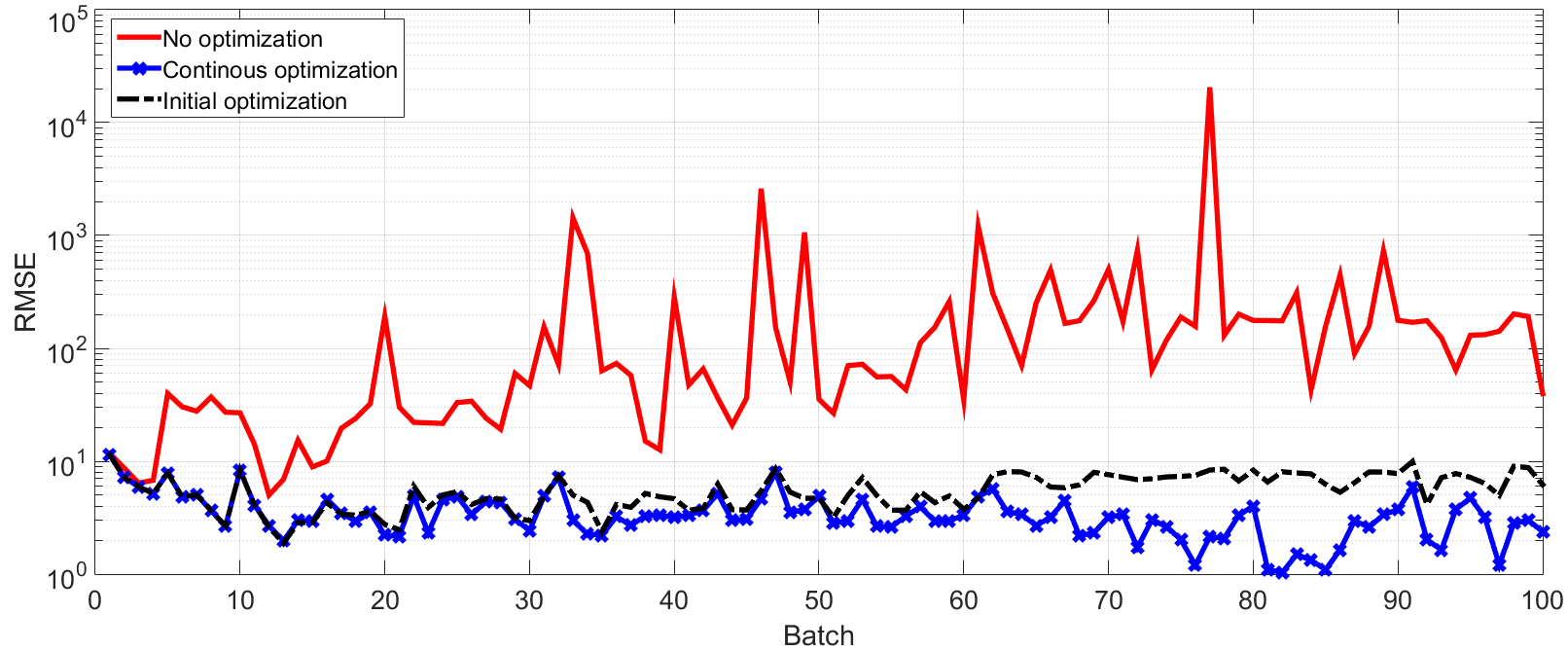}
\caption{Accuracy for the Sarcos dataset using the polynomial kernel with hyperparameter optimization and without Random Matrix Factorization}
\label{fig:REMSE_Sarcos_polyk}
\end{figure}

We now demonstrate the hyperparameter optimization method listed in Algorithm~\ref{algo:seqGPhyper}, where a simple quadratic kernel from Section~\ref{sec:with-optimization} is employed. There is no need for any manual hyperparameter selection for this technique. In order to quantify the importance and the overhead of automatically tuning the hyperparameters, we implemented three scenarios: no-optimization, continuous optimization and initial optimization. 

Figures~\ref{fig:REMSE_Abalone_polyk} and~\ref{fig:REMSE_Sarcos_polyk} show the accuracy (measured by RMSE and log(RMSE)) for the Abalone and Sarcos datasets, respectively. For both datasets, when no optimization is applied, the error keeps accumulating over the iterations. For the continuous optimization mode,  the hyperparameters are optimized at each step (when a new batch is received). It is shown that the accuracy is acceptable (similar to that obtained by employing the less restrictive  exponential kernel and by manually tuning the hyperparameters  (Tables \ref{table:expKNRMF} and \ref{table:polyKNRMF}). However, the computational overhead of optimizing the hyperparameters makes the continuous optimization mode infeasible. To address this trade-off between efficiency and accuracy, we implemented the initial optimization mode, where we optimize the hyperparameters over a fixed number of initial steps, and thereafter we keep the hyperparameters constant for the subsequent steps. It is clear from the figures that the accuracy is reasonable when we let the algorithm optimize the hyperparameters for only ten steps.

We observe that it is not feasible to measure the actual accuracy improvement over the iterations since the testing set is different at each specific iteration (we only predict for samples in the new arriving batch at every iteration). However, it is clear from the plots that the error does not increase/accumulate over the iterations even though more noise is added to the data over the iterations since the predictions (not the actual values) update the GP model at each iteration.

\begin{figure}[!t]
\centering
\includegraphics[width=\columnwidth]{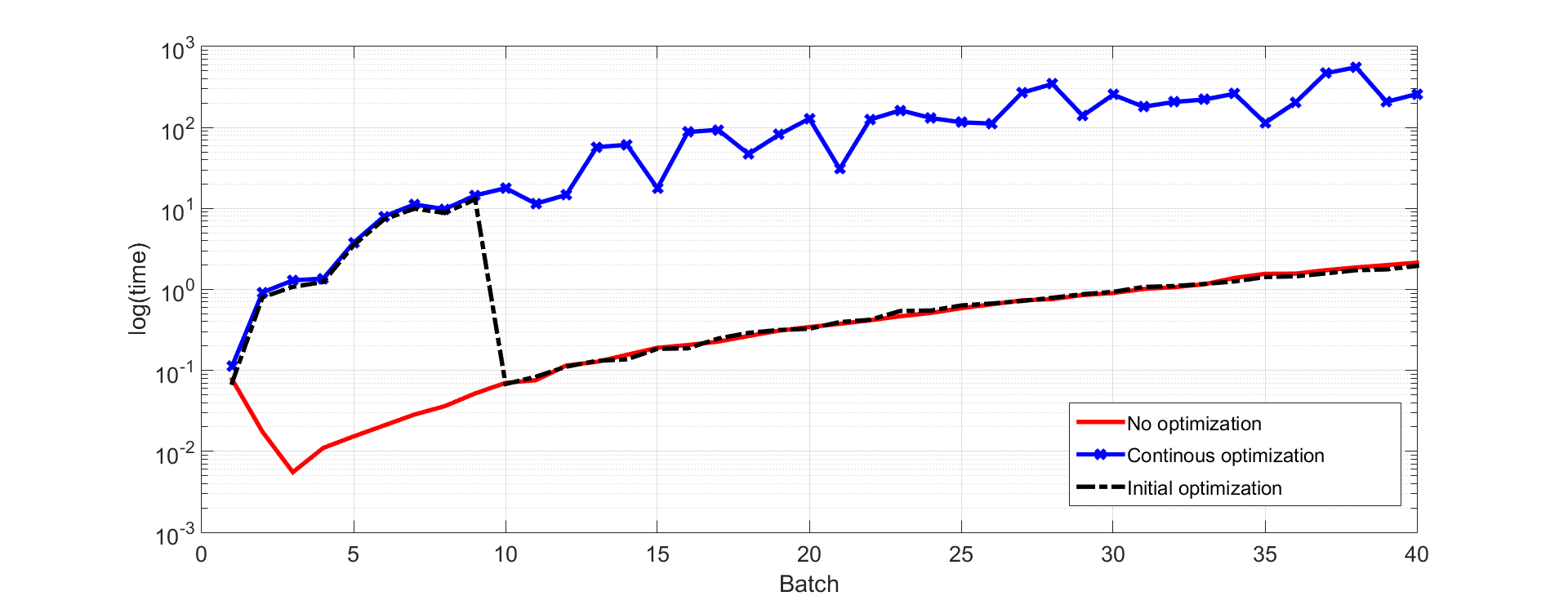}
\caption{Computational time for the Abalone dataset using the polynomial kernel with different hyperparameter optimization modes and without Random Matrix Factorization}
\label{fig:time_Abalone_polyk_NRMF}
\end{figure}

\begin{figure}[!t]
\centering
\includegraphics[width=\columnwidth]{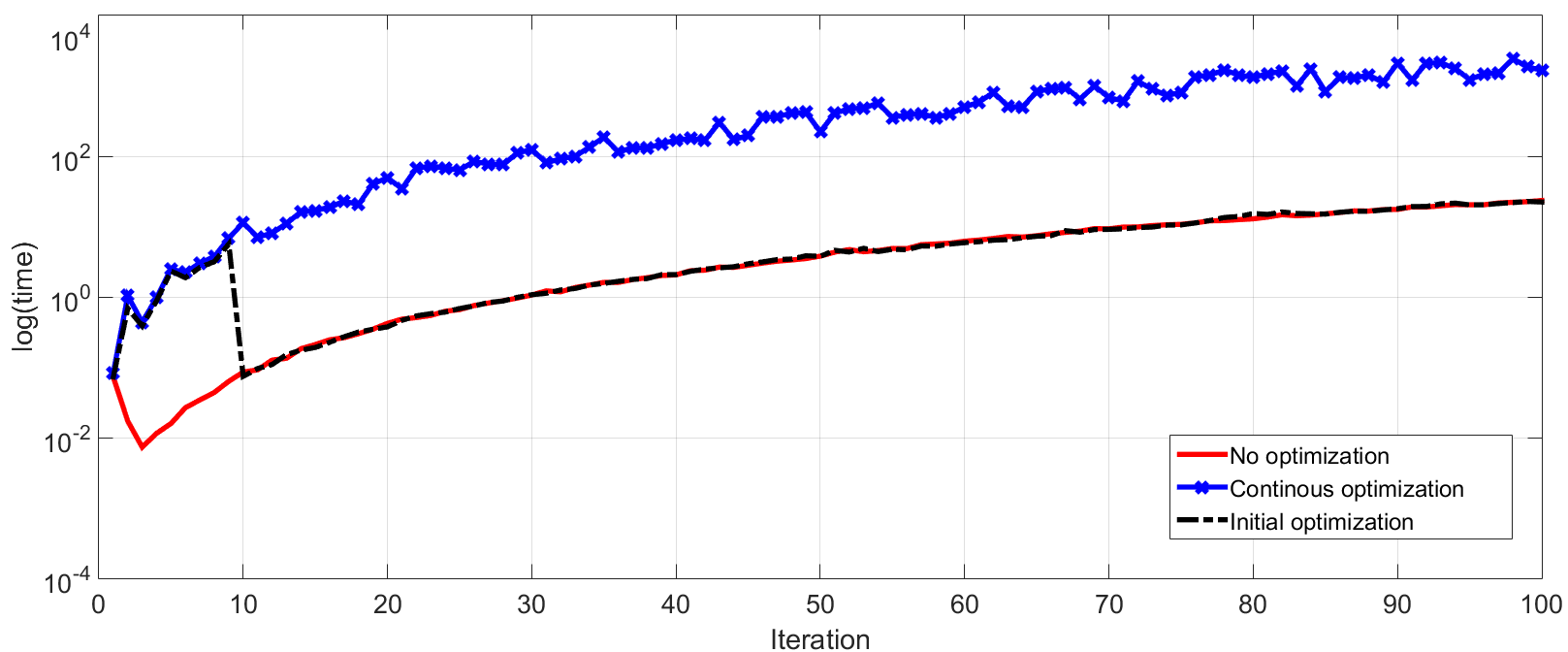}
\caption{Computational time for the Sarcos dataset using the polynomial kernel with different hyperparameter optimization modes and without Random Matrix Factorization}
\label{fig:time_Sarcos_polyk_NRMF}
\end{figure}

Figures~\ref{fig:time_Abalone_polyk_NRMF} and~\ref{fig:time_Sarcos_polyk_NRMF} show that employing the initial optimization mode dropped the computational cost to same level as when no optimization step is applied.   Accordingly, we focus only on the initial optimization mode for the rest of this paper.

Table \ref{table:polyKNRMF} summarizes the results of the three aforementioned optimization modes. It is clear that running the incremental GP without tuning the hyperparameters results into very large errors. On the other hand, continuously optimizing the hyperparameters is computationally infeasible. Updating the hyperparameters during the first initial steps results in a reasonable compromise of efficiency and accuracy, where the computational gain is much higher than the accuracy loss. Although the quadratic kernel is more restrictive than the squared exponential kernel, it provides comparable results for both datasets, when the hyperparameters are optimized using our technique. More specifically, for the Sarcos dataset, when the squared exponential kernel is manually tuned (see Table \ref{table:expKNRMF}, NRMF algorithm), the average prediction time per batch was similar to that when a polynomial kernel is automatically tuned (see Table \ref{table:polyKNRMF}, initial optimization mode) while the accuracy is increased. For the Abalone dataset, the optimization method is less effective, however, it is important to note that the quadratic kernel is more restrictive as well. 


So far, we observe that it is more promising to automatically update the hyperparameters than manually tuning them upfront. However, it is still computationally expensive to optimize the hyperparameters without applying random matrix factorization. More specifically, the average batch time is 1.72 sec and 6.68 sec, for the Abalone and Sarcos datasets, respectively (see Table \ref{table:polyKNRMF}, initial optimization mode).  Also, it has been shown, in Section \ref{sec:Without Hyperparameter Optimization}, that employing the proposed sequential random matrix factorization (SRMF) approach increases the efficiency of the manually optimized streaming Gaussian processes. More specifically, by employing the SRMF algorithm, the average batch time is decreased from 0.63 sec and 6.68 to 0.21 sec and 1.15 sec, for the Abalone and Sarcos datasets, respectively (see Table
\ref{table:expKNRMF}).


Accordingly, it is promising to employ the matrix factorization approach with the automatic \emph{hybrid} parameter optimization scenario (as described in Section \ref{sec:with-optimization} and Algorithm \ref{algo:seqGPhyper}). Since we observe that initially optimizing the hyperparameters provides a reasonable compromise between accuracy and efficiency, we now test a variant of Algorithm~\ref{algo:seqGPhyper} where hyperparameters are optimized during the first ten steps, and then they are fixed for the remaining dataset. Since there is no need to optimize the parameters after the initialization steps, we employ the distance matrix factorization only for the initialization steps, and then the kernel matrix factorization (as listed in algorithm \ref{algo:seqGP}) is applied for the subsequent steps. When the initial optimization mode is applicable, employing the hybrid approach becomes  more efficient than Algorithm \ref{algo:seqGPhyper} since factoring the kernel matrix is faster and possibly more robust than factoring the distance matrix.

\begin{table}[!t]
\caption{Comparison of Algorithms for Incremental GP with Polynomial Kernel AND  With Hyperparameter Optimization}
\centering
\begin{tabular}{|c|c|c|c|c|}
\hline
\multirow{2}{*}{Algorithm}& \multicolumn{2}{|c|}{Abalone}  & \multicolumn{2}{|c|}{Sarcos}\\\cline{2-5}
 {}& RMSE & Time & RMSE& TIME\\
\hline
NRMF& 5.06 & 1.72 &5.66& 6.86\\
\hline
BRMF&  5.47 & 1.42&6.93& 4.08\\
\hline
SRMF& 4.76 & 0.86&5.42&1.81 \\
\hline
\end{tabular}
\label{table:hyperparameters}
\end{table}

\begin{figure}[!t]
\centering
\includegraphics[width=\columnwidth]{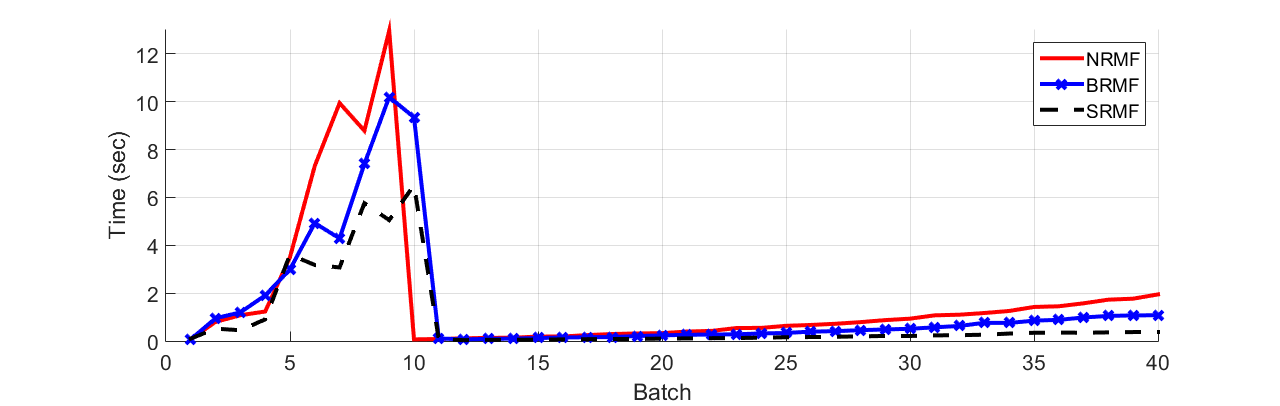}
\caption{Scalability for the Abalone dataset using the polynomial kernel with hyperparameter optimization and with Random Matrix Factorization}
\label{fig:time_Abalone_polyk}
\end{figure}

\begin{figure}[!t]
\centering
\includegraphics[width=\columnwidth]{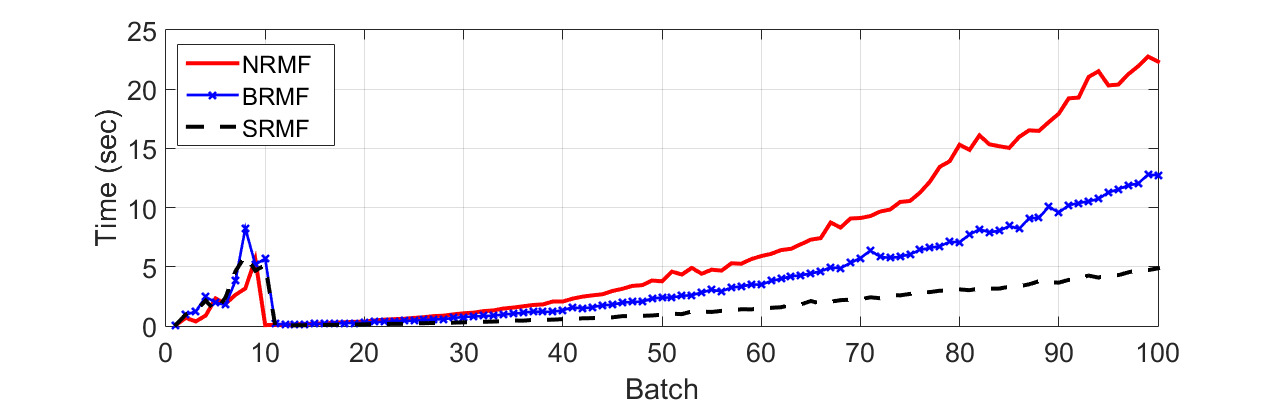}
\caption{Scalability for the Sarcos dataset using the polynomial kernel with hyperparameter optimization and with Random Matrix Factorization}
\label{fig:time_Sarcos_polyk}
\end{figure}

Figures~\ref{fig:time_Abalone_polyk} and~\ref{fig:time_Sarcos_polyk} show the computational time  for the Abalone and Sarcos datasets, respectively. For both datasets, applying the random matrix factorization approaches increased the efficiency significantly, where the proposed SRMF algorithm outperforms the other algorithms. Table \ref{table:hyperparameters} summarizes the results for employing  the hyperparameter optimization approach at the initial steps with the quadratic kernel.  It is clear that the proposed method significantly decreased the computation cost while provided a slightly enhanced accuracy compared to the NRMF approach.

\section{Conclusion and Future Directions}\label{sec:conclusion}
This paper formalized a sequential technique based on randomized low-rank matrix factorization approach for incrementally predicting values of an unknown function at test points using the Gaussian Processes framework. The use of the factored form of the kernel matrix serves as an efficient way to store increasing amounts of data points, as well as provides an advantage in computing the predicted values. In terms of accuracy, the proposed method yields higher error, which is still within acceptable limit compared to the batch approach on publicly available real datasets. Compared to a state of the art recursive technique, the proposed approach yields a comparable error and is computationally more efficient up to a certain data size. Additionally, we also introduced a new kernel function that leads to an efficient technique to update the hyperparameters.

An immediate future direction is to extend the technique to more general kernel functions. Other directions include developing hybrid techniques based on a combination of the proposed approach and the approach from \cite{huber2014} and exploring distributed aspects of computation in streaming scenarios.

\section*{Acknowledgment}

This work was supported by the Office of Naval Research under the contract N00014-16-C-2009.

\ifCLASSOPTIONcaptionsoff
  \newpage
\fi



%

\bibliographystyle{ieeetr}
\end{document}